\documentclass{article}
\usepackage{nips07submit_e,times}
\usepackage{nips07submit_e,algorithm,algorithmic,times,changebar,wrapfig,subfigure,trackchanges,url,xspace,epsfig,mydefs,verbatim}

\title{The Infinite Hierarchical Factor Regression Model}

\vspace{-4em} 
\author{
Piyush Rai and Hal Daum\'e III\\
School of Computing, University of Utah\\
\texttt{\{piyush,hal\}@cs.utah.edu}
}

\begin{document} 
 
\maketitle 

%\vspace{-2.5em} 
\begin{abstract} 
  We propose a nonparametric Bayesian factor regression model that accounts for uncertainty in the number of factors, and the relationship between factors.  To accomplish this, we propose a sparse variant of the Indian Buffet Process and couple this with a hierarchical model over factors, based on Kingman's coalescent. We apply this model to two problems (factor analysis and factor regression) in gene-expression data analysis.
\end{abstract} 
  
\section{Introduction} 

Factor analysis is the task of explaining data by means of a set of
\emph{latent factors}.  Factor \emph{regression} couples this analysis
with a prediction task, where the predictions are made solely on the
basis of the factor representation. The latent factor representation achieves two-fold
benefits: (1) discovering the latent \textit{process} underlying the data; (2) simpler predictive modeling
through a compact data representation. In particular, (2) is motivated by the problem of prediction in the \textit{``large P small N''} paradigm \cite{bfrm}, where the number of features $P$ greatly exceeds the number of examples $N$, potentially resulting in overfitting. 

We address three fundamental
shortcomings of standard factor analysis approaches
\cite{sabattijameshidden,sanglawrratregact,bealrecoreg,bfrm}: (1) we
do not assume a known number of factors; (2) we do not assume factors
are independent; (3) we do not assume all features are relevant to the
factor analysis.  Our motivation for this work stems from the task of
reconstructing regulatory structure from gene-expression data.  In
this context, factors correspond to regulatory pathways.  Our
contributions thus parallel the needs of gene pathway modeling. In addition, we couple predictive modeling (for factor regression) within the factor analysis framework itself, instead of having to model it separately.

Our factor regression model is fundamentally nonparametric.  In
particular, we treat the gene-to-factor relationship nonparametrically
by proposing a sparse variant of the Indian Buffet Process (IBP)
\cite{ibp}, designed to account for the sparsity of relevant genes
(features).  We \emph{couple} this IBP with a hierarchical prior over
the factors.  This prior explains the fact that pathways are
fundamentally related: some are involved in transcription, some in
signaling, some in synthesis.  The nonparametric nature of our sparse IBP requires that the hierarchical prior \emph{also} be nonparametric.
A natural choice is Kingman's coalescent \cite{kingcoal}, a popular
distribution over infinite binary trees.

Since our motivation is an application in bioinformatics, our
notation and terminology will be drawn from that area.  In particular,
\emph{genes} are \emph{features}, \emph{samples} are
\emph{examples}, and \emph{pathways} are \emph{factors.}  However,
our model is more general.  An alternative application might be to a
collaborative filtering problem, in which case our genes might
correspond to movies, our samples might correspond to users and our
pathways might correspond to genres.  In this context, all three
contributions of our model still make sense: we do not know how many
movie genres there are; some genres are closely related (romance to
comedy versus to action); many movies may be spurious.

%\vspace{-1em} 
\section{Background}
\label{sec:background}
%\vspace{-0.5em} 

Our model uses a variant of the Indian Buffet Process to model the
feature-factor (i.e., gene-pathway) relationships.  We further use Kingman's coalescent to
model latent pathway hierarchies.

%\vspace{-0.5em}  
\subsection{Indian Buffet Process}
\label{sec:ibp}
%\vspace{-0.5em} 

The Indian Buffet Process \cite{ibp1} defines a distribution over
infinite binary matrices, originally motivated by the need to model
the latent factor structure of a given set of observations.  In the
standard form it is parameterized by a scale value, $\alpha$.  The
distribution can be explained by means of a simple culinary analogy.
Customers (in our context, \emph{genes}) enter an Indian restaurant
and select dishes (in our context, \emph{pathways}) from an infinite
array of dishes. The first customer selects $Poisson(\alpha)$
dishes. Thereafter, each incoming customer $i$ selects a
previously-selected dish $k$ with a probability $m_k/(i-1)$, where
$m_k$ is the number of previous customers who have selected dish $k$.
Customer $i$ then selects an \emph{additional} $Poisson(\alpha/i)$ new
dishes.  We can easily define a binary matrix $\bf{Z}$ with value
$Z_{ik}=1$ precisely when customer $i$ selects dish $k$.  This
stochastic process thus defines a distribution over infinite binary
matrices. 

It turn out \cite{ibp1} that the stochastic process defined above
corresponds to an infinite limit of an exchangeable process over
finite matrices with $K$ columns.  This distribution takes the form
$p(\textbf{Z} \| \alpha) = \prod_{k=1}^K
\frac{\frac{\alpha}{K}\Gamma(m_k + \frac{\alpha}{K})\Gamma(P - m_k
  -1)}{\Gamma(P + 1 + \frac{\alpha}{K})}$, where $m_k =
\sum_{i}Z_{ik}$ and $P$ is the total number of customers.  Taking $K \rightarrow \infty$ yields the IBP.  The
IBP has several nice properties, the most important of which is
exchangeablility.  It is the exchangeablility (over samples) that
makes efficient sampling algorithms possible. There also exists a two-parameter generalization to IBP where the second parameter $\beta$ controls the sharability of dishes.
\vspace{-0.5em}
\subsection{Kingman's Coalescent}

Our model makes use of a latent hierarchical structure over factors;
we use Kingman's coalescent \cite{kingcoal} as a convenient prior
distribution over hierarchies.  Kingman's coalescent originated in the
study of population genetics for a set of single-parent organisms.
The coalescent is a nonparametric model over a countable set of
organisms.  It is most easily understood in terms of its finite
dimensional marginal distributions over $n$ individuals, in which case
it is called an $n$-coalescent.  We then take the limit $n \rightarrow
\infty$.  In our case, the individuals are \emph{factors}.

The $n$-coalescent considers a population of $n$ organisms at time
$t=0$.  We follow the ancestry of these individuals backward in time,
where each organism has exactly one parent at time $t<0$.  The
$n$-coalescent is a continuous-time, partition-valued Markov process
which starts with $n$ singleton clusters at time $t=0$ and evolves
\emph{backward}, coalescing lineages until there is only one left.  We
denote by $t_i$ the \emph{time} at which the $i$th coalescent event
occurs (note $t_i \leq 0$), and $\de_i = t_{i-1} - t_{i}$ the time
between events (note $\de_i > 0$).  Under the $n$-coalescent, each
pair of lineages merges indepentently with exponential rate $1$; so
$\de_i \sim \Exp\left({n-i+1}\choose{2}\right)$.  With probability
one, a random draw from the $n$-coalescent is a binary tree with a
single root at $t=-\infty$ and $n$ individuals at time $t=0$.  We
denote the tree structure by $\pi$.  The marginal distribution over
tree topologies is uniform and independent of coalescent times; and
the model is infinitely exchangeable.  We therefore consider the limit
as $n \rightarrow \infty$, called \emph{the coalescent.}

Once the tree structure is obtained, one can define an additional
Markov process to evolve over the tree.  One common choice is a
Brownian diffusion process.  In Brownian diffusion in $D$ dimensions,
we assume an underlying diffusion covariance of $\vec \La \in \R^{D
  \times D}$ p.s.d.  The root is a $D$-dimensional vector drawn $\vec
z$.  Each non-root node in the tree is drawn Gaussian with mean equal
to the value of the parent, and variance $\de_i \vec \La$, where
$\de_i$ is the time that has passed.

Recently, Teh et al. \cite{agglocoal} proposed efficient bottom-up
agglomerative inference algorithms for the coalescent.  These
(approximately) maximize the probability of $\pi$ and $\de$s,
marginalizing out internal nodes by Belief Propagation.  If we
associate with each node in the tree a \emph{mean} $\vec y$ and
\emph{variance} $\vec v$ message, we update messages as
Eq~\eqref{eq:bp}, where $i$ is the current node and $li$ and $ri$ are
its children.
\begin{align}
\vec v_{i} &= \left[
(\vec v_{{li}}+(t_{li}-t_i)\vec \La)\inv +
(\vec v_{{ri}}+(t_{ri}-t_i)\vec \La)\inv\right]\inv
\label{eq:bp}\\
\vec y_{i} &= \left[
\vec y_{li} (\vec v_{{li}}+(t_{li}-t_i)\vec \La)\inv +
\vec y_{ri} (\vec v_{{ri}}+(t_{ri}-t_i)\vec \La)\inv\right]\inv
\vec v_{i}
\nonumber
\end{align} 

\section{Nonparametric Bayesian Factor Regression}
\label{sec:npbfrm}
\vspace{-0.5em}
Recall the standard factor analysis
problem: $\textbf{X} = \textbf{AF} + \textbf{E}$, for standardized
data \textbf{X}. \textbf{X} is a $P \times N$ matrix consisting of $N$
samples [$\vec{x}_1,...,\vec{x}_N$] of $P$ features each. \textbf{A}
is the factor loading matrix of size $P \times K$ and \textbf{F} =
[$\vec{f}_1,...,\vec{f}_N$] is the factor matrix of size $K \times
N$. \textbf{E} = [$\vec{e}_1,...,\vec{e}_N$] is the matrix of
idiosyncratic variations. $K$, the number of factors, is known.

Recall that our goal is to treat the factor analysis problem
nonparametrically, to model feature relevance, and to model
hierarchical factors.  For expository purposes, it is simplest to deal
with each of these issues in turn. In our context, we begin by modeling the
gene-factor relationship nonparametrically (using the IBP).  Next, we
propose a variant of IBP to model gene relevance.  We then
present the hierarchical model for inferring factor hierarchies.  We conclude with a presentation of the full model and our mechanism for modifying the factor
\emph{analysis} problem to factor \emph{regression}.
\vspace{-0.5em}
\subsection{Nonparametric Gene-Factor Model}
\vspace{-0.5em}
We begin by directly using the IBP to infer the number of
factors. Although IBP has been applied to nonparametric factor
analysis in the past \cite{ibp}, the standard IBP formulation places
IBP prior on the factor matrix ($\bf{F}$) associating \emph{samples}
(i.e. a set of features) with factors. Such a model assumes that
the sample-fctor relationship is sparse. However, this assumption is
inappropriate in the gene-expression context where it is not the
factors themselves but the \emph{associations} among genes and factors (i.e., the factor loading matrix $\bf{A}$) that are sparse. In such a context, each sample depends on all the factors but each gene within a sample usually depends only on a small number of factors.

Thus, it is more appropriate to model the factor loading matrix
($\bf{A}$) with the IBP prior. %In the corresponding culinary analogy
%of IBP, the customers are \emph{genes} which are selecting dishes
%(factors). 
Note that since \textbf{A} and \textbf{F} are related with
each other via the number of factors \textsl{K}, modeling \textbf{A}
nonparametrically allows our model to also have an unbounded number of
factors.%, essentially leading to an alternative nonparametric latent
%factor model.

\begin{comment} Another limitation of the IBP model for factor
  analysis is that it assumes binary valued factors which is not a
  reasonable assumption for many realistic scenarios.\end{comment}

%\subsection{Nonparametric Factor Analysis}

For most gene-expression problems \cite{bfrm}, a binary factor
loadings matrix ($\bf{A}$) is inappropriate.  Therefore, we instead
use the Hadamard (element-wise) product of a binary matrix \textbf{Z}
and a matrix \textbf{V} of reals. \textbf{Z} and \textbf{V} are of the
same size as \textbf{A}. The factor analysis model, for each sample
$i$, thus becomes: $ \vec{x}_i = (\vec{Z} \odot \vec{V})\vec{f}_i +
\vec{e}_i$. We have $\textbf{Z} \sim \IBP(\alpha,\beta)$. $\alpha$ and
$\beta$ are IBP hyperparameters and have vague gamma priors on them.  Our
initial model assumes no factor hierarchies and hence the prior over \textbf{V} would simply be a Gaussian: $\textbf{V} \sim \Nor(0,\sigma_v^2\bf{I})$ with an inverse-gamma prior on $\sigma_v$. $\bf{F}$ has a zero mean, unit variance Gaussian prior, as used in
standard factor analysis. Finally, $\vec{e}_i = \Nor(0,\bf{\Psi})$
models the idiosyncratic variations of genes where $\bf{\Psi}$ is a $P
\times P$ diagonal matrix ($diag(\Psi_1,...,\Psi_P)$). Each entry
$\Psi_P$ has an inverse-gamma prior on it. 
%Posterior inference is done
%by suitably modifying the Gibbs sampler for IBP to also propose new
%columns of $\vec{V}$ matrix whenever new columns of $\vec{Z}$ matrix
%are proposed. The inference details are given in section
%\ref{sec:infer}.

\begin{comment}or $\textbf{V} \sim Coalescent()$\footnote{The coalescent prior is
  defined as $\Nor(\pi|\pi_{\mu},\pi_{\Sigma})$}, if we assume factor
hierarchy. The details of coalescent prior and related inference
procedure for that case are described in section
\ref{sec:coalv}. \end{comment}

\subsection{Feature Selection Prior}
\label{sec:varsel}
%\vspace{-0.5em} 

Typical gene-expression datasets are of the order of several thousands
of genes, most of which are \emph{not} associated with any pathway
(factor).  In the above, these are accounted for only by the
idiosyncratic noise term.  A more realistic model is that certain
genes simply do not participate in the factor analysis: for a culinary
analogy, the genes enter the restaurant and leave before selecting any
dishes.  Those genes that ``leave'', we term ``spurious.''  We add an
additional prior term to account for such spurious genes; effectively
leading to a sparse solution (over the rows of the IBP matrix).  It is important to note
that this notion of sparsity is fundamentally \emph{different} from
the conventional notion of sparsity in the IBP.  The sparsity in IBP
is over \emph{columns}, not \emph{rows.}  To see the difference,
recall that the IBP contains a ``rich get richer'' phenomenon:
frequently selected factors are more likely to get reselected.
Consider a truly spurious gene and ask whether it is likely to select
any factors.  If some factor $k$ is already frequently used, then
\emph{a priori} this gene is more likely to select it.  The only
downside to selecting it is the data likelihood.  By setting the
corresponding value in $\bf{V}$ to zero, there is no penalty.

Our sparse-IBP prior is identical to the standard IBP prior with one
exception.  Each customer (gene) $p$ is associated with Bernoulli
random variable $T_p$ that indicates whether it samples \emph{any}
dishes.  The $\bf{T}$ vector is given a parameter $\rho$, which, in
turn, is given a Beta prior with parameters $a,b$.

\subsection{Hierarchical Factor Model}
\label{sec:coalv}

In our basic model, each column of the matrix $\vec{Z}$ (and the
corresponding column in $\vec{V}$) is associated with a factor.  These
factors are considered unrelated.  To model the fact that factors are,
in fact, related, we introduce a factor hierarchy.  Kingman's
coalescent \cite{kingcoal} is an attractive prior for integration with
IBP for several reasons. It is nonparametric and describes
exchangeable distributions.  This means that it can model a varying
number of factors.  Moreover, efficient inference algorithms exist
\cite{agglocoal}.  

\begin{figure}[ht]
\begin{minipage}[b]{0.48\linewidth}
\centering
\includegraphics[scale=0.20]{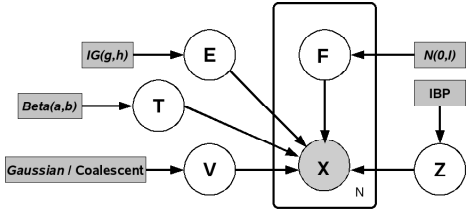}
\caption{\small{The graphical model for nonparametric Bayesian Factor Regression. \textbf{X} consists of response variables as well.}}
\label{fig:gm}
\end{minipage}
\begin{minipage}[b]{0.48\linewidth}
\centering
\includegraphics[scale=0.40]{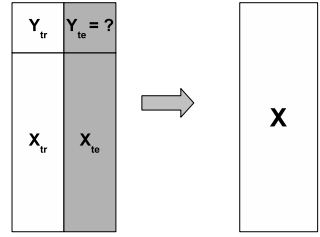}
\caption{\small{Training and test data are combined together and test responses are treated as missing values to be imputed}}
\label{fig:lumped}
\end{minipage}
\end{figure}

\subsection{Full Model and Extension to Factor Regression}
\label{sec:regression}
Our proposed graphical model is depicted in Figure~\ref{fig:gm}.  The
key aspects of this model are: the IBP prior over $\bf{Z}$, the sparse
binary vector $\bf{T}$, and the Coalescent prior over $\bf{V}$.

In standard Bayesian factor regression \cite{bfrm}, factor analysis is
followed by the regression task.  The regression is performed only on
the basis of $\bf{F}$, rather than the full data
$\bf{X}$.  For example, a simple linear regression problem would
involve estimating a $K$-dimensional parameter vector $\vec \theta$
with regression value $\vec \th\T \bf{F}$.  Our model, on the other
hand, integrates factor regression component in the nonparametric
factor analysis framework itself. We do so by prepending the responses
$y_i$ to the expression vector $\vec{x}_i$ and joining the training
and test data (see figure \ref{fig:lumped}). The unknown responses in
the test data are treated as missing variables to be iteratively
imputed in our MCMC inference procedure. It is straightforward to see
that it is equivalent to fitting another sparse model relating factors to responses. Our model thus 
allows the factor analysis to take into account the regression task as well.
In case of binary responses, we add an extra probit regression step
to predict binary outcomes from real-valued responses.

%\vspace{-1.1em}
\section{Inference}
\label{sec:infer}
%\vspace{-1.1em}
We use Gibbs sampling with a few M-H steps.  The Gibbs
distributions are summarized here.

\textbf{Sampling the IBP matrix Z:} Sampling \textbf{Z} consists of
sampling existing dishes, proposing new dishes and accepting or
rejecting them based on the acceptance ratio in the associated M-H
step. For sampling existing dishes, an entry in $\bf{Z}$ is set as 1
according to
$p(Z_{ik}=1|\textbf{X},Z_{-ik},\textbf{V},\textbf{F},\mathbf{\Psi})
\propto \frac{m_{-i,k}}{(P + \beta -
  1)}p(\textbf{X}|\textbf{Z},\textbf{V},\textbf{F},\mathbf{\Psi})$
whereas it is set as 0 according to
$p(Z_{ik}=0|\textbf{X},Z_{-ik},\textbf{V},\textbf{F},\mathbf{\Psi})
\propto \frac{P + \beta - 1 - m_{-i,k}}{(P + \beta -
  1)}p(\textbf{X}|\textbf{Z},\textbf{V},\textbf{F},\mathbf{\Psi})$. $m_{-i,k}
= \sum_{j \neq i}Z_{jk}$ is how many other customers chose dish $k$.

For sampling new dishes, we use an M-H step where we simultaneously
propose $\mathbf{\eta} = (K^{new}, V^{new}, F^{new})$ where $K^{new}
\sim Poisson(\alpha\beta/(\beta+P-1))$. We accept the proposal with an
acceptance probability (following \cite{dyadibp}) given by $a =
\min\{1,\frac{p(rest|\mathbf{\eta^*})}{p(rest|\mathbf{\eta})}\}$. Here,
$p(rest|\eta)$ is the likelihood of the data given parameters
$\eta$. We propose $V^{new}$ from its prior (either Gaussian or
Coalescent) but, for faster mixing, we propose $F^{new}$ from its
posterior.

\begin{wrapfigure}{r}{1.2in}
\vspace{-2em}
\includegraphics[width=1.1in]{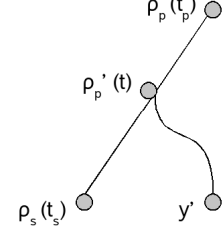}
\caption{Adding a new node to the tree}
%\vspace{-1em}
\label{fig:coaltree}
\end{wrapfigure}

Sampling $V^{new}$ from the coalescent is slightly involved.  As shown
pictorially in figure \ref{fig:coaltree}, proposing a new column of
$\vec{V}$ corresponds to adding a new leaf node to the existing
coalescent tree. In particular, we need to find a sibling ($s$) to the
new node $y'$ and need to find an insertion point on the branch
joining the sibling $s$ to its parent $p$ (the grandparent of $y'$).
Since the marginal distribution over trees under the coalescent is
uniform, the sibling $s$ is chosen uniformly over nodes in the tree.
We then use importance sampling to select an insertion time for the
new node $y'$ between $t_s$ and $t_p$, according to the exponential
distribution given by the coalescent prior (our proposal distribution
is uniform). This gives an insertion point in the tree, which
corresponds to the new parent of $y'$. We denote this new parent by
$p'$ and the time of insertion as $t$. The predictive density of the
newly inserted node $y'$ can be obtained by marginalizing the parent
$p'$.  This yields $\Nor(\vec y_0, \vec v_0)$, given by:
\begin{align} \label{eq:predictive-likelihood}
\vec v_0 &= [(\vec v_{s}+(t_s-t)\vec \La)\inv + (\vec v_{p}+(t-t_p)\vec \La)\inv]\inv
\nonumber\\
\vec y_0 &= [ \vec y_{s} / (v_{s}+(t_s-t)\vec \La) + \vec y_{p} / (v_{p}+(t_p-t)\vec \La) ]\vec v_0  \nonumber
\end{align}
Here, $y_{s}$ and $v_{s}$ are the messages passed \emph{up} through
the tree, while $y_{p}$ and $v_{p}$ are the messages passed
\emph{down} through the tree (compare to Eq~\eqref{eq:bp}).

\textbf{Sampling the sparse IBP vector T:} In the \emph{sparse IBP
  prior}, recall that we have an additional $P$-many variables $T_p$,
indicating whether gene $p$ ``eats'' any dishes.  $T_p$ is drawn from
Bernoulli with parameter $\rho$, which, in turn, is given a
$\Bet(a,b)$ prior.  For inference, we collapse $\rho$ and
$\mathbf{\Psi}$ and get Gibbs posterior over $T_p$ of the form
${p(T_p=1|.)} \propto (a + \sum_{q \neq
  p}T_p)\Stu(\vec{x}_p|(\textbf{Z}_p \odot \textbf{V}_p)\vec{F},g/h,g))$
and $p(T_p=0|.) \propto (b + P - \sum_{q \neq
  p}T_q)\Stu(\vec{x}_p|0,g/h,g)$, where $\Stu$ is the non-standard
Student's t-distribution. $g,h$ are hyperparameters of the
inverse-gamma prior on the entries of $\mathbf{\Psi}$.

\textbf{Sampling the real valued matrix V:} For the case when
\textbf{V} has a Gaussian prior on it, we sample \textbf{V} from its
posterior $
p(V_{g,j}|\textbf{X},\textbf{Z},\textbf{F},\mathbf{\Psi}) \propto
\Nor(V_{g,j}|\mu_{g,j},\Sigma_{g,j})$, where $\Sigma_{g,j} =
(\sum_{i=1}^N\frac{F_{j,i}^2}{\Psi_g} + \frac{1}{\sigma_v^2})^{-1}$
and $\mu_{g,j} = \Sigma_{g,j}(\sum_{i=1}^N
F_{j,i}X^*_{g,j})\Psi_g^{-1}$. We define $X^*_{g,j} = X_{g,i} -
\sum_{l=1,l \neq j}^K (A_{g,l}V_{g,l})F_{l,i}$, and $\textbf{A} =
\textbf{Z} \odot \textbf{V}$. The hyperparameter $\sigma_v$ on
\textbf{V} has an inverse-gamma prior and posterior also has the same
form.  For the case with coalescent prior on \textbf{V}, we have
$\Sigma_{g,j} = (\sum_{i=1}^N\frac{F_{j,i}^2}{\Psi_g} +
\frac{1}{{v_0}_j})^{-1}$ and $\mu_{g,j} = \Sigma_{g,j}(\sum_{i=1}^N
F_{j,i}X^*_{g,j})(\Psi_g + \frac{{y_0}_{g,j}}{{v_0}_j})^{-1}$, where
$\vec y_0$ and $\vec v_0$ are the Gaussian posteriors of the leaf node added in
the coalescent tree (see Eq~\eqref{eq:bp}), which corresponds to the column of \textbf{V} being sampled.

\textbf{Sampling the factor matrix F:} We sample for \textbf{F} from its posterior $p(\textbf{F}|\textbf{X},\textbf{Z},\textbf{V},\mathbf{\Psi}) \propto \Nor(\textbf{F} |\mathbf{\mu},\mathbf{\Sigma})$ where $\bf{\mu} = \bf{A}^T(\bf{A}\bf{A}^T + \Psi)^{-1}X$ and $\bf{\Sigma} = \textbf{I} - \bf{A}^T(\bf{A}\bf{A}^T + \Psi)^{-1}\bf{A}$, where $\textbf{A} = \textbf{Z} \odot \textbf{V}$

\textbf{Sampling the idiosyncratic noise term:} We place an inverse-gamma prior on the diagonal entries of $\mathbf{\Psi}$ and the posterior too is inverse-gamma: $p(\Psi_p|.) \propto \IG(g + \frac{N}{2},\frac{h}{1+\frac{h}{2}tr(\textbf{E}^T\textbf{E})})$, where $\textbf{E} = \textbf{X} - \bf{(Z \odot V)F}$.

\textbf{Sampling IBP parameters:} We sample the IBP parameter $\alpha$ from its posterior: $p(\alpha|.) \sim \Gam(K_+ + a,\frac{b}{1 + bH_P(\beta)})$,
where $K_+$ is the number of active features at any moment and $H_P(\beta) = \sum_{i=1}^P 1/(\beta + i -1)$. $\beta$ is sampled from a prior proposal using an M-H step.

\textbf{Sampling the Factor Tree:} Use the \textsf{Greedy-Rate1} algorithm
\cite{agglocoal}.
\section{Related Work} \label{sec:priorwork}
\vspace{-0.5em}
A number of probabilistic approaches have been proposed in the past for the problem of gene-regulatory network reconstruction \cite{sabattijameshidden,sanglawrratregact,bealrecoreg,bfrm}.  Some
take into account the information on the prior network topology
\cite{sabattijameshidden}, which is not always available.  Most assume
the number of factors is known.  To get around this, one can perform
model selection via Reversible Jump MCMC \cite{green95reversible} or
evolutionary stochastic model search \cite{sparsest}. Unfortunately,
these methods are often difficult to design and may take quite long to
converge.  Moreover, they are difficult to integrate with other forms
of prior knowledge (eg., factor hierarchies). A somewhat similar approach to ours is the infinite independent component analysis (iICA) model of \cite{knowlesica} which treats factor analysis as a special case of ICA. However, their model is limited to factor analysis and does not take into account feature selection, factor hierarchy and factor regression. As a generalization to the standard ICA model, \cite{tca} proposed a model in which the components can be related via a tree-structured graphical model. It, however, assumes a fixed number of components.

Structurally, our model with Gaussian-$\textbf{V}$ (i.e. no hierarchy over factors) is most similar to
the Bayesian Factor Regression Model (BFRM) of \cite{bfrm}.  BFRM assumes a sparsity inducing mixture
prior on the factor loading matrix $\textbf{A}$. Specifically, $A_{pk}
\sim (1 - \pi_{pk})\delta_0(A_{pk}) + \pi_{pk} \Nor(A_{pk}|0,\tau_k)$
where $\delta_0()$ is a point mass centered at zero. To complete the
model specification, they define $\pi_{pk} \sim (1 -
\rho_k)\delta_0(\pi_{pk}) + \rho_k \Bet(\pi_{pk}|sr,s(1-r))$ and
$\rho_k \sim \Bet(\rho_k|av,a(1-v))$. Now, integrating out $\pi_{pk}$
gives: $A_{pk} \sim (1-v\rho_k)\delta_0(A_{pk}) + v\rho_k
\Nor(A_{pk}|0,\tau_k)$.  It is interesting to note that the
nonparametric prior of our model (factor loading matrix defined as
$\textbf{A} = \bf{Z \odot V}$) is actually equivalent to the
(parametric) sparse mixture prior of the BFRM as
$K\!\rightarrow\!\infty$.  To see this, note that our prior on
the factor loading matrix $\textbf{A}$ (composed of $\textbf{Z}$
having an IBP prior, and $\textbf{V}$ having a Gaussian prior), can be
written as $A_{pk} \sim (1- \rho_k)\delta_0(A_{pk}) + \rho_k
\Nor(A_{pk}|0,\sigma_v^2)$, if we define $\rho_k \sim
\Bet(1,\alpha\beta/K)$. It is easy to see that, for BFRM where $\rho_k
\sim \Bet(av,a(1-v))$, setting $a = 1+\alpha\beta/K$ and
$v=1-\alpha\beta/(aK)$ recovers our model in the limiting case when
$K\!\rightarrow\!\infty$.
\vspace{-0.5em}
\section{Experiments} 
\label{sec:exper}
\vspace{-0.5em}
In this section, we report our results on synthetic and real datasets. We compare our nonparametric approach with the evolutionary search based approach proposed in \cite{sparsest}, which is the nonparametric extension to BFRM.

We used the gene-factor connectivity matrix of E-coli network (described in \cite{pourwernbmc}) to generate a synthetic dataset having 100 samples of 50 genes and 8 underlying factors. Since we knew the ground truth for factor loadings in this case, this dataset was ideal to test for efficacy in recovering the factor loadings (binding sites and number of factors). We also experimented with a real gene-expression data which is a breast cancer dataset having 251 samples of 226 genes and 5 prominent underlying factors (we know this from domain knowledge).
\vspace{-0.5em}
\subsection{Nonparametric Gene-Factor Modeling and Variable Selection}
\begin{figure}[!htbp]
\begin{center}
\begin{tabular}{ccc}
\includegraphics[width=1.8in]{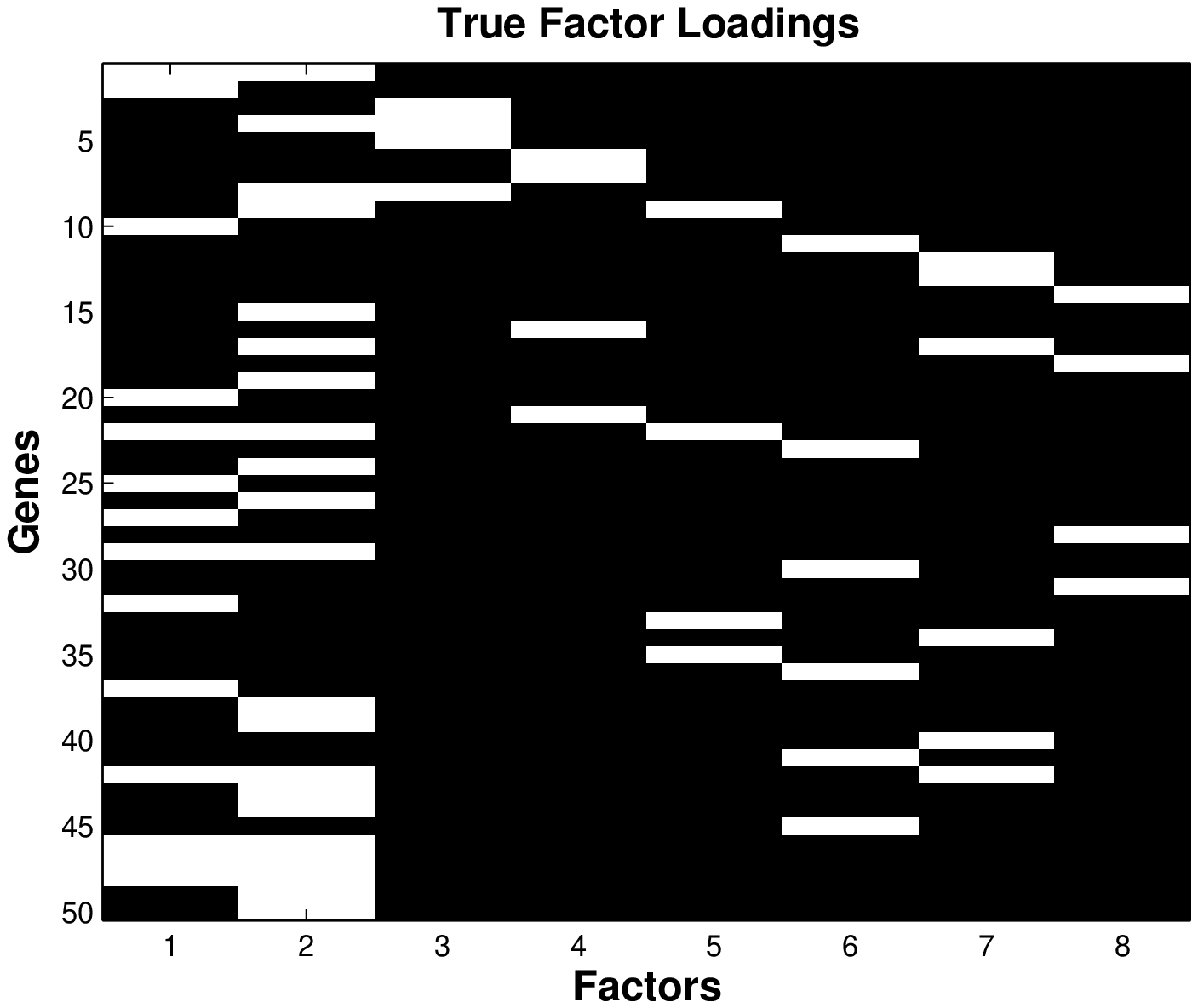}
\includegraphics[width=1.8in]{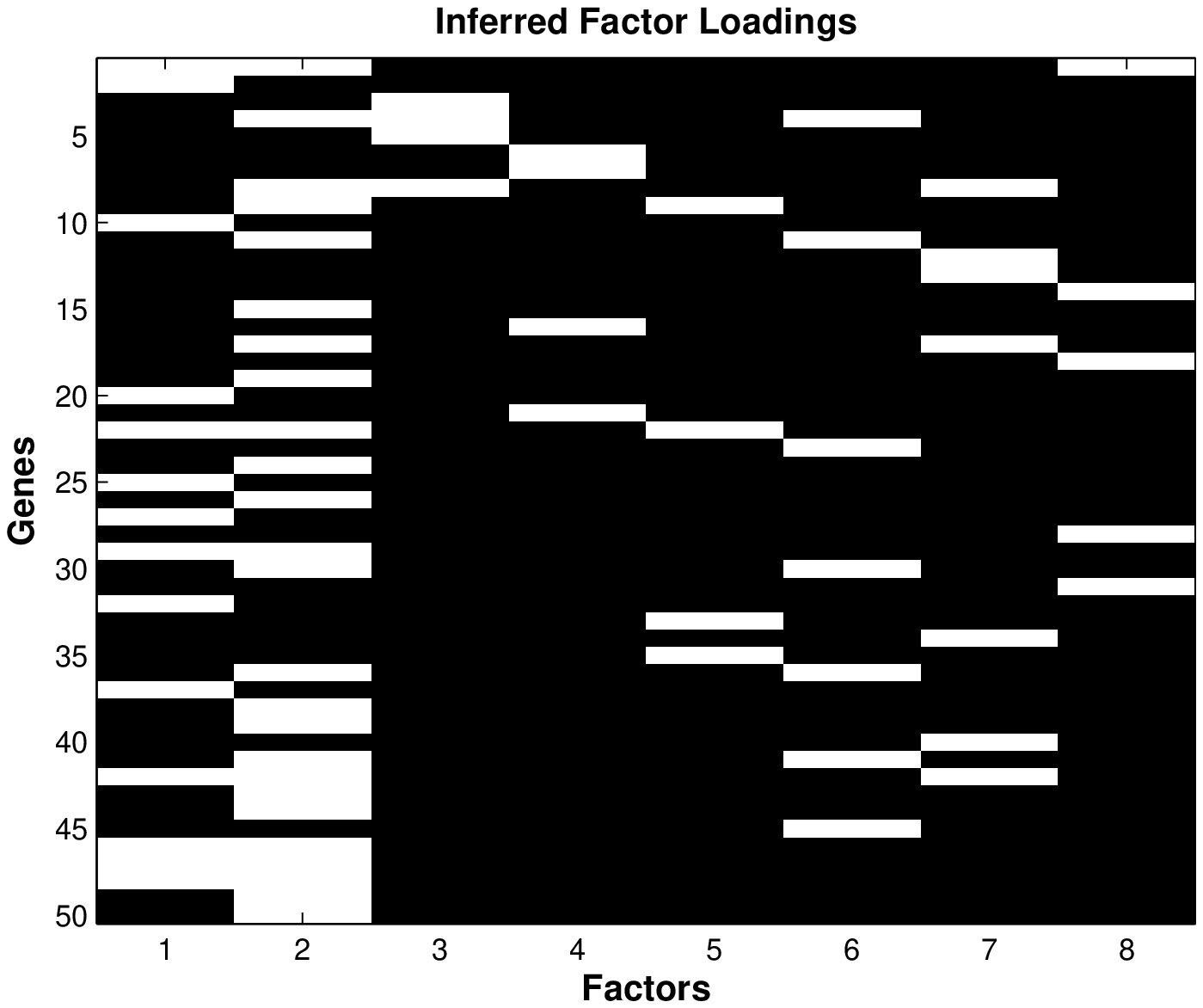}
\includegraphics[width=1.8in]{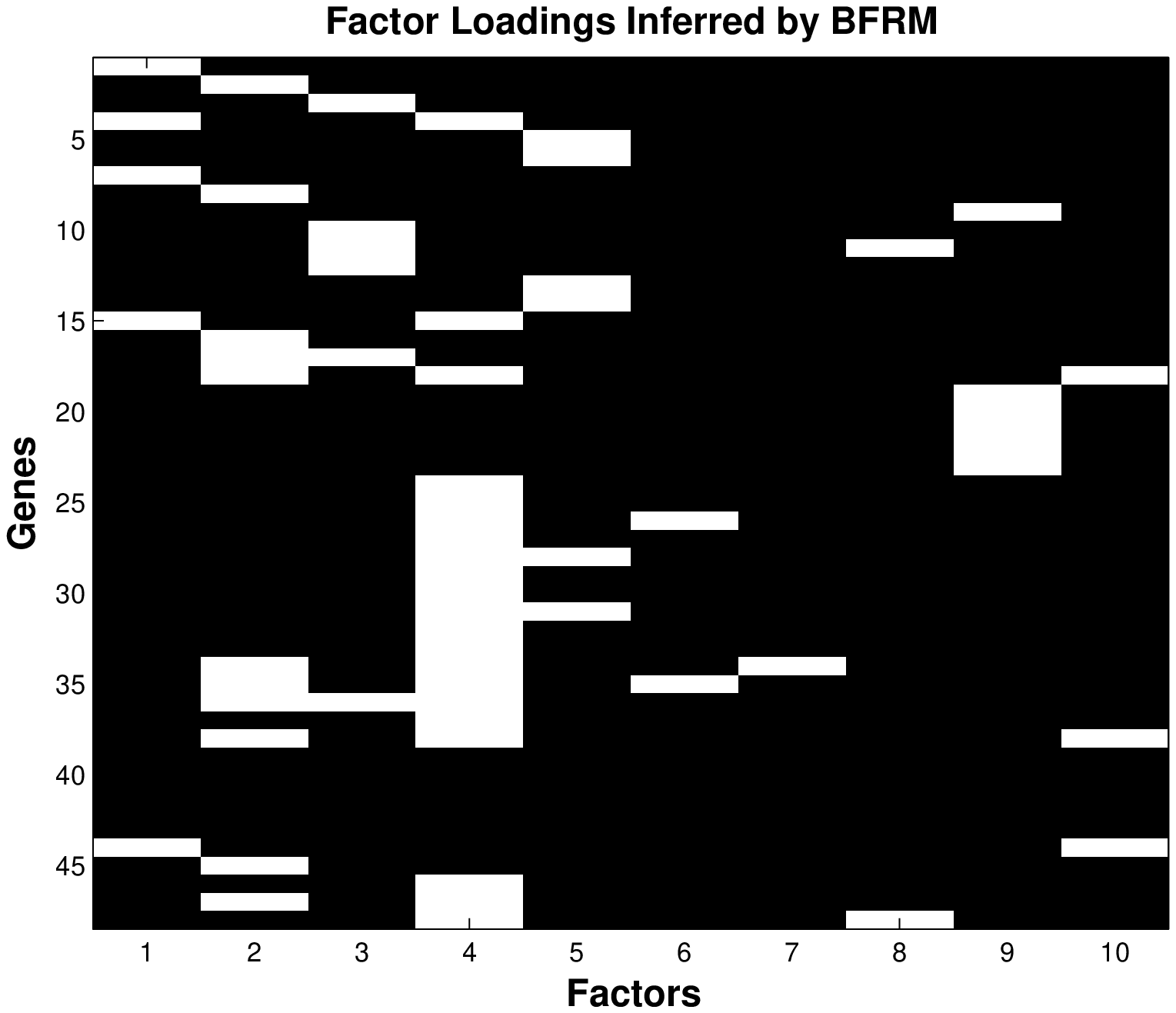}
\end{tabular}
\end{center}
\vspace{-1.5em}
\caption{\small{(Left and middle) True and inferred factor loadings (with our approach) for the synthetic data with P=50, K=8 generated using connectivity matrix of E-coli data. (Right) Inferred factor loadings with the evolutionary search based approach. White rectangles represent active sites. The data also has added noise with signal-to-noise-ratio of 10}}
\label{fig:p50}
\end{figure}

%\textbf{Synthetic Data:} 
For the synthetic dataset generated by the E-coli network, the results are shown in figure \ref{fig:p50} comparing the actual network used to generate the data and the inferred factor loading matrix. As shown in figure \ref{fig:p50}, we recovered exactly the same number (8) of factors, and almost exactly the same factor loadings (binding sites and number of factors) as the ground truth. In comparison, the evolutionary search based approach overestimated the number of factors and the inferred loadings clearly seem to be off from the actual loadings (even modulo column permutations).

\vspace{-1.0em}

\begin{figure*}[!htbp]
   \centering
   \subfigure[] {
       \includegraphics[width=1.9in]{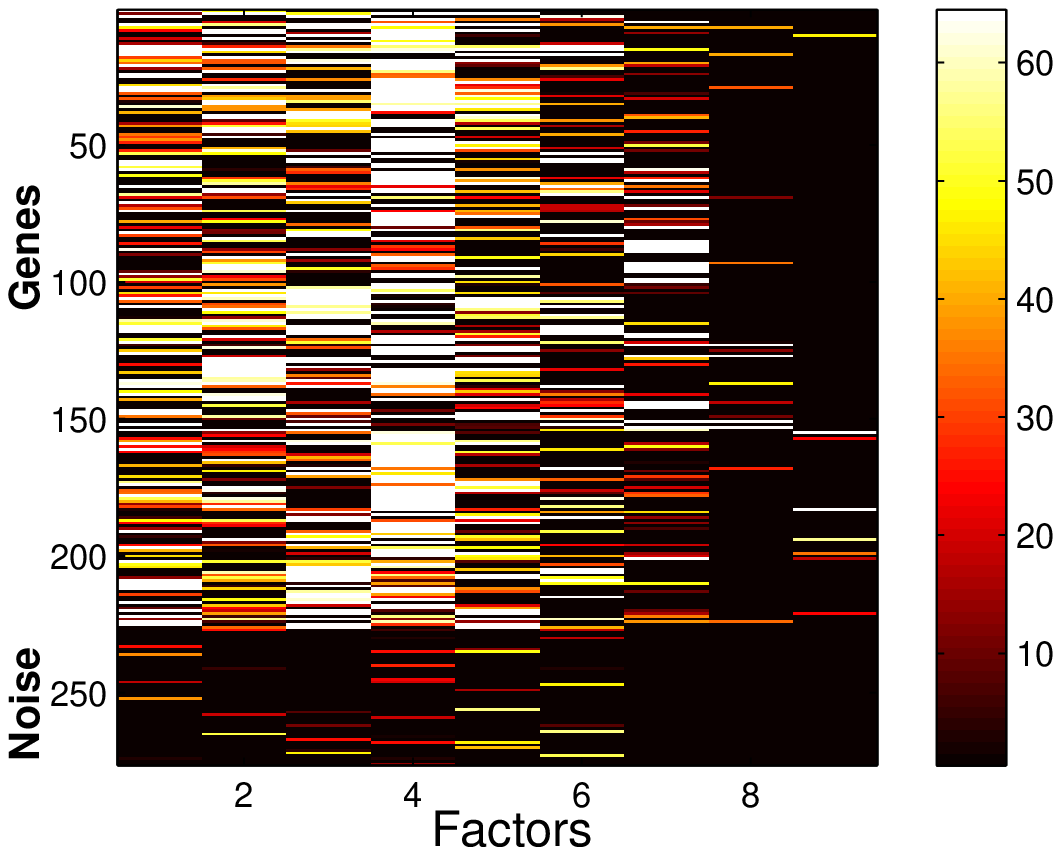}
       \label{fig:spwofsel}
   }
   \hspace{-2em}
   \subfigure[] {
       \includegraphics[width=1.9in]{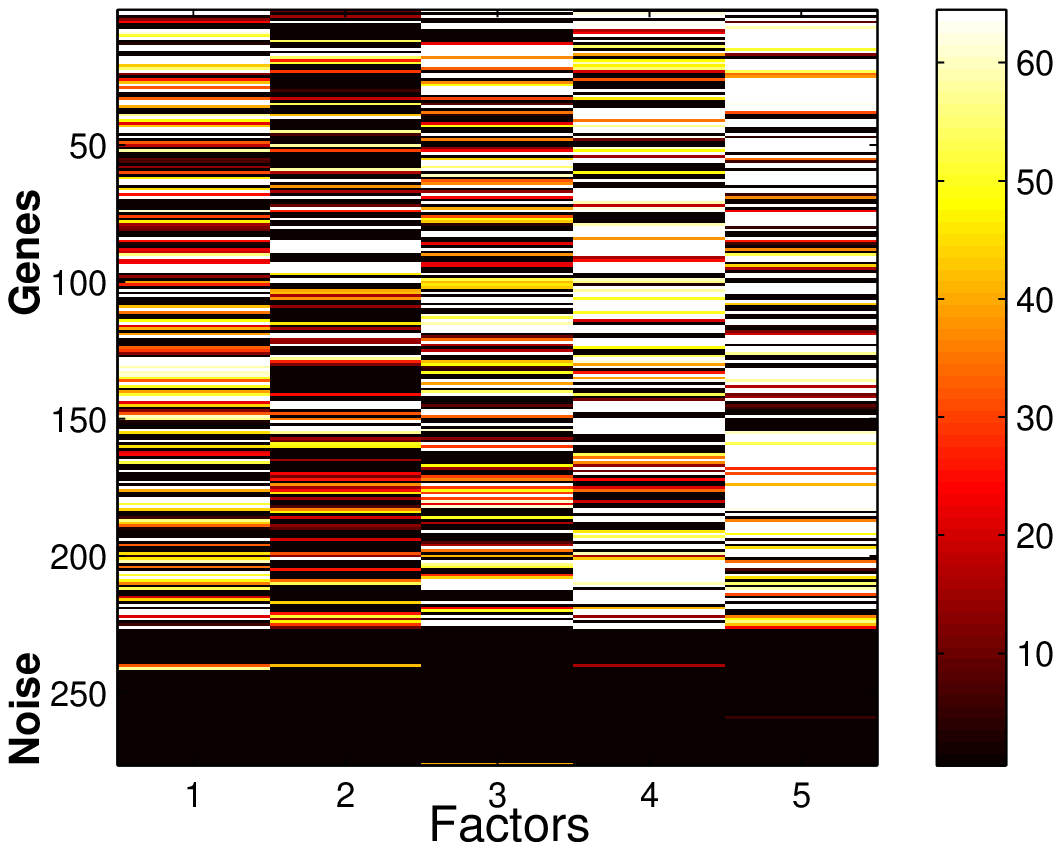}
       \label{fig:spwfsel}
   }
   \hspace{-2em}
   \subfigure[] {
       \includegraphics[width=1.7in]{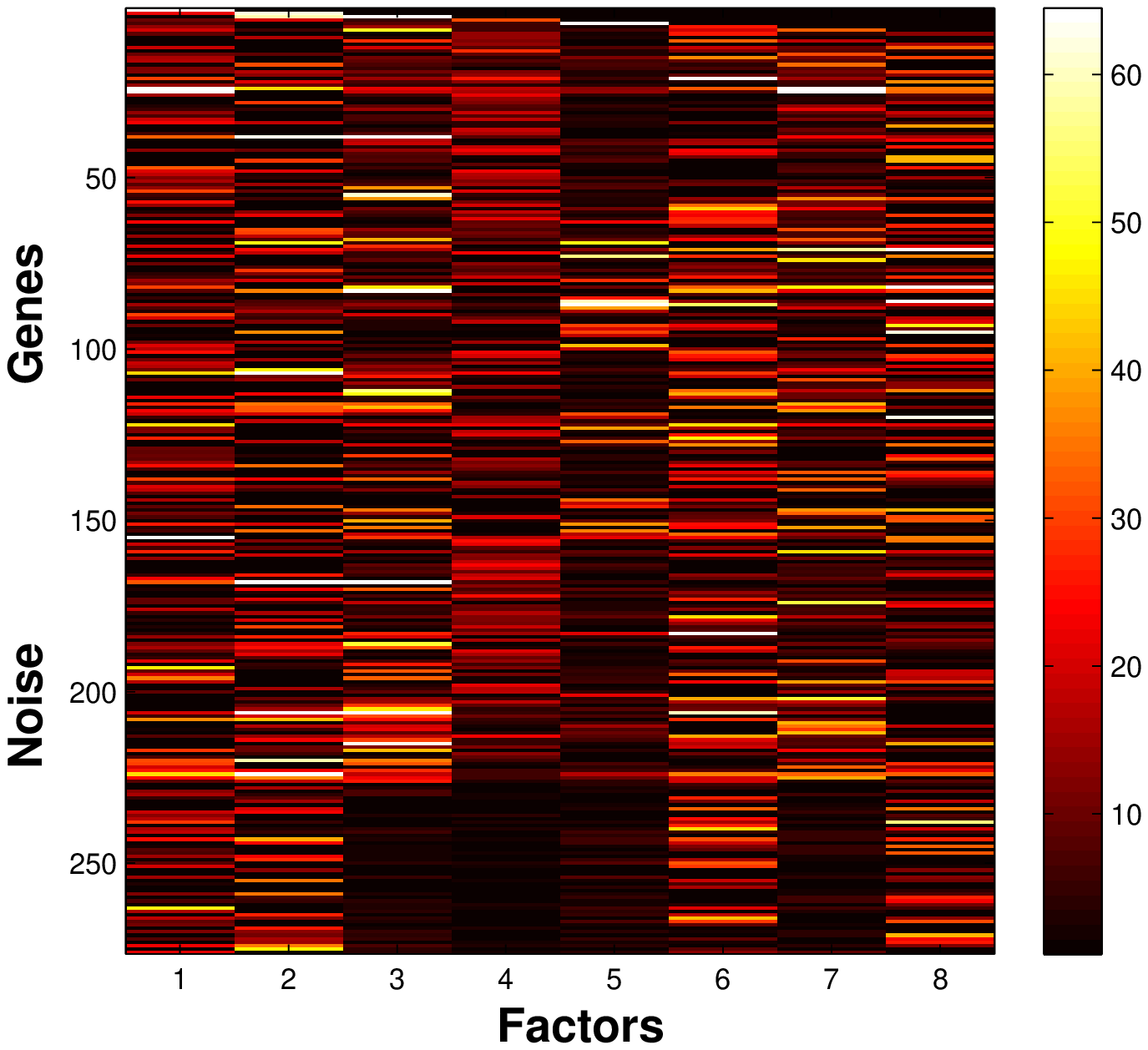}
       \label{fig:spbfrm}
   }      
   \caption{\small{Effect of spurious genes (heat-plots of factor loading matrix shown): (a) Standard IBP (b) Our model with variable selection (c) The evolutionary search based approach}}
   \label{fig:realdata}
\end{figure*}

%\textbf{Real Data:} 
Our results on real data are shown in figure \ref{fig:realdata}. To see the effect of variable selection for this data, we also introduced spurious genes by adding 50 random features in each sample. We observe the following: (1) Without variable selection being on, spurious genes result in an overestimated number of factors and falsely discovered factor loadings for spurious genes (see figure \ref{fig:spwofsel}), (2) Variable selection, when on, effectively filters out spurious genes, without overestimating the number of factors (see figure \ref{fig:spwfsel}). We also investigated the effect of noise on the evolutionary search based approach and it resulted in an overestimated number of factor, plus false discovered factor loadings for spurious genes (see figure \ref{fig:spbfrm}). To conserve space, we do not show here the cases when there are no spurious genes in the data but it turns out that variable selection does not filter out any of 226 relevant genes in such a case.

\vspace{-1.0em}

\subsection{Hierarchical Factor Modeling} 
\begin{figure*}[!htbp]
   \centering
   \subfigure[] {
       \includegraphics[width=1.4in]{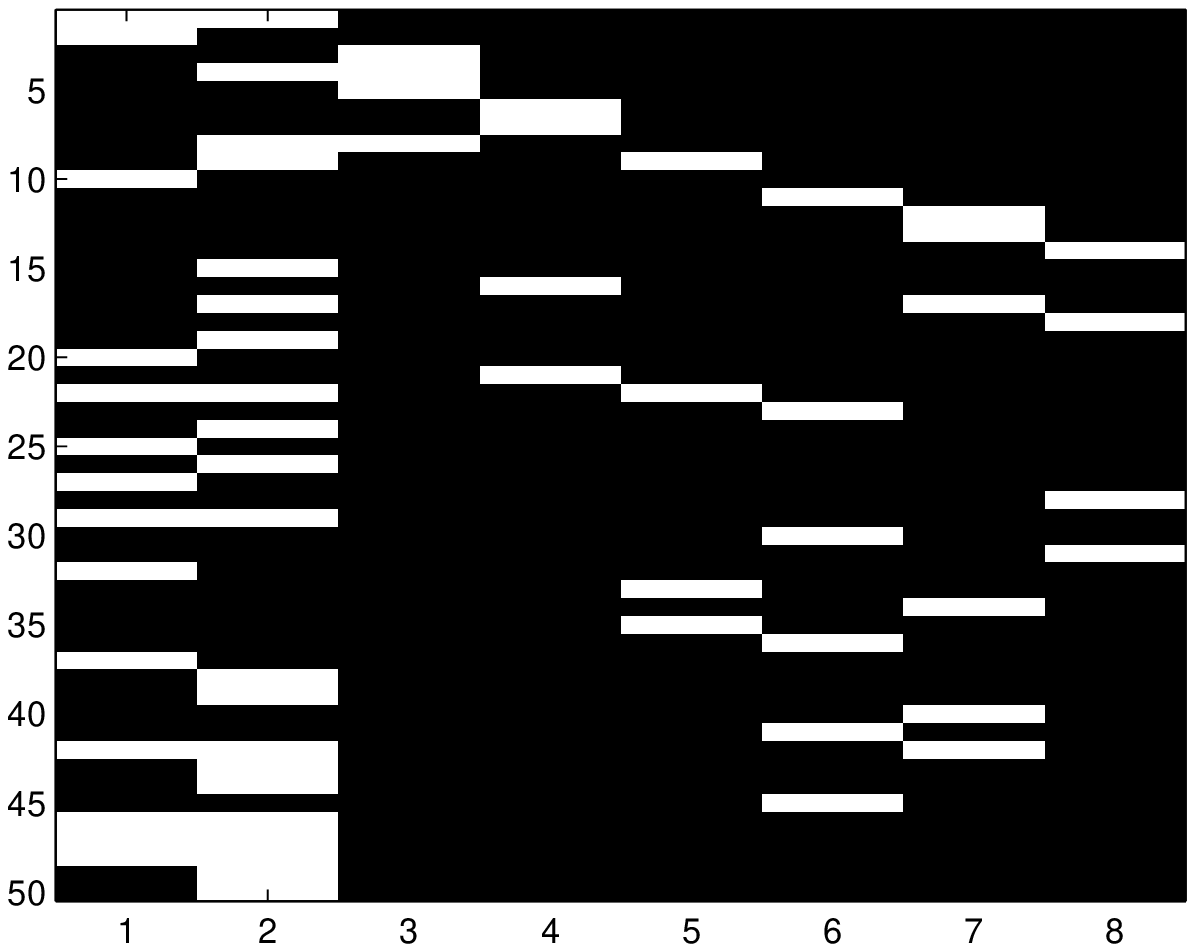}
       \label{fig:}
   }
\hspace{-1em}
   \subfigure[] {
       \includegraphics[width=1.4in]{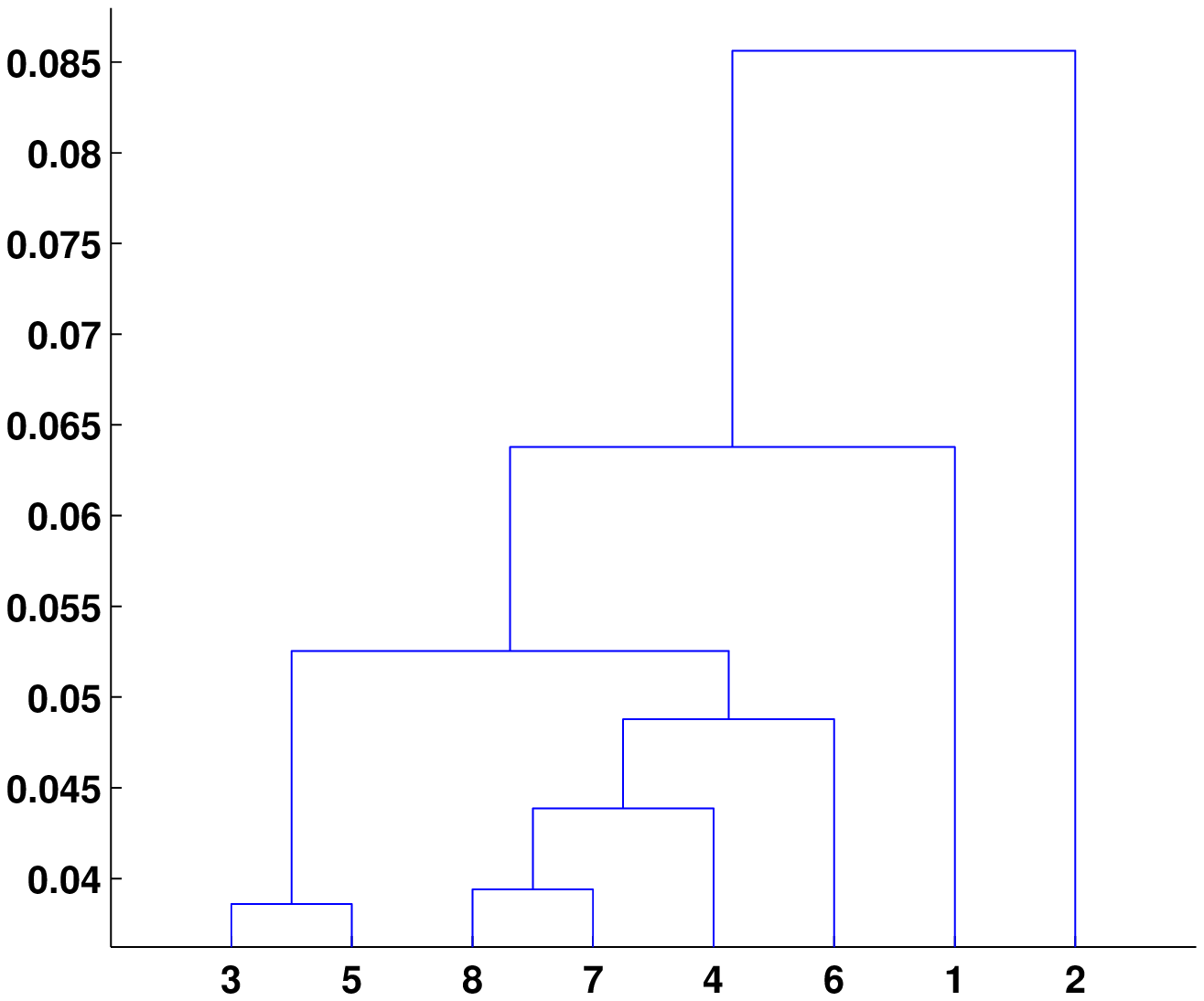}
       \label{fig:}
   }
\hspace{-1em} 
   \subfigure[] {
       \includegraphics[width=1.4in]{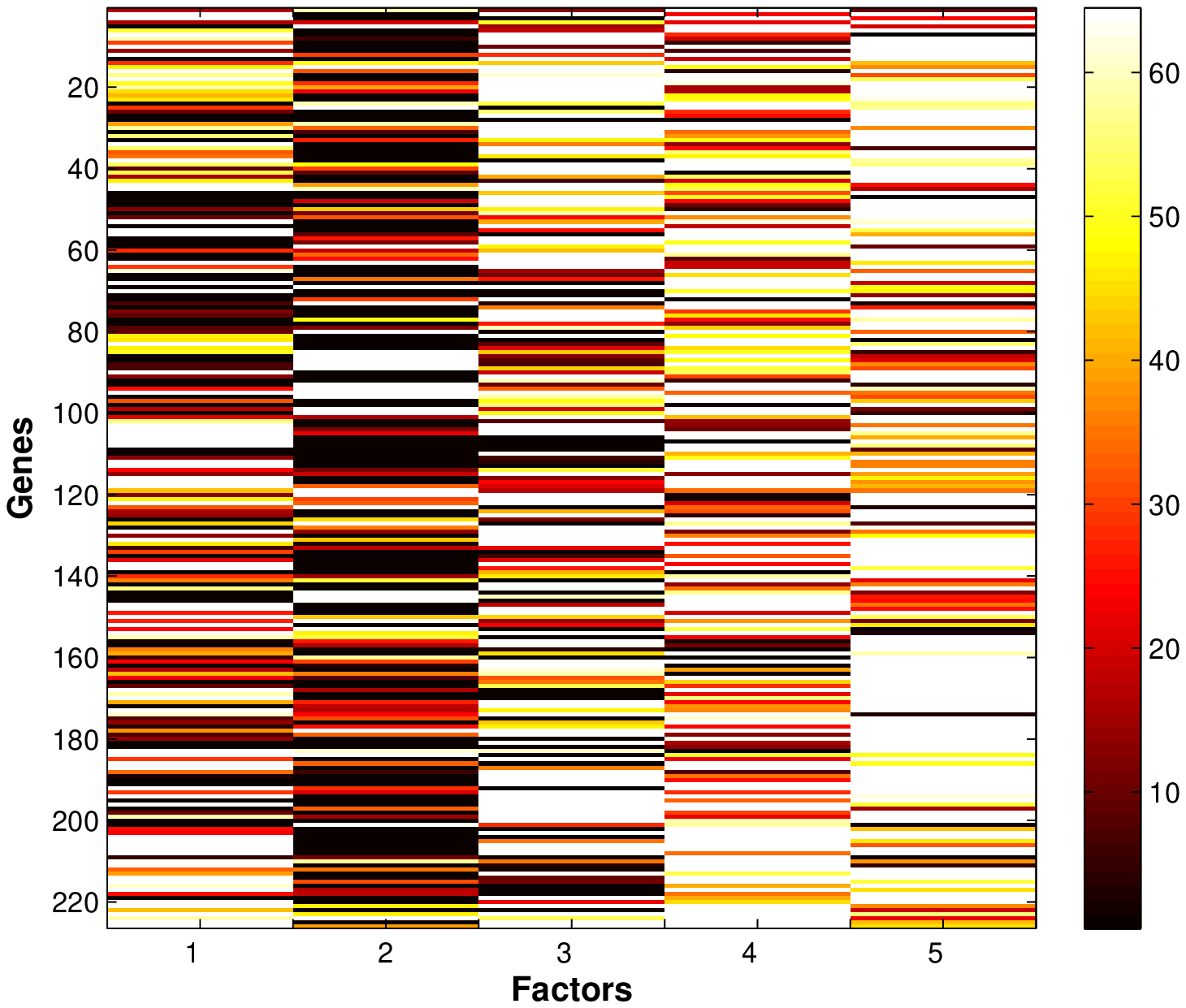}
       \label{fig:}
   }
\hspace{-0.5em}
   \hspace{-2em}
   \subfigure[] {
       \includegraphics[width=1.4in]{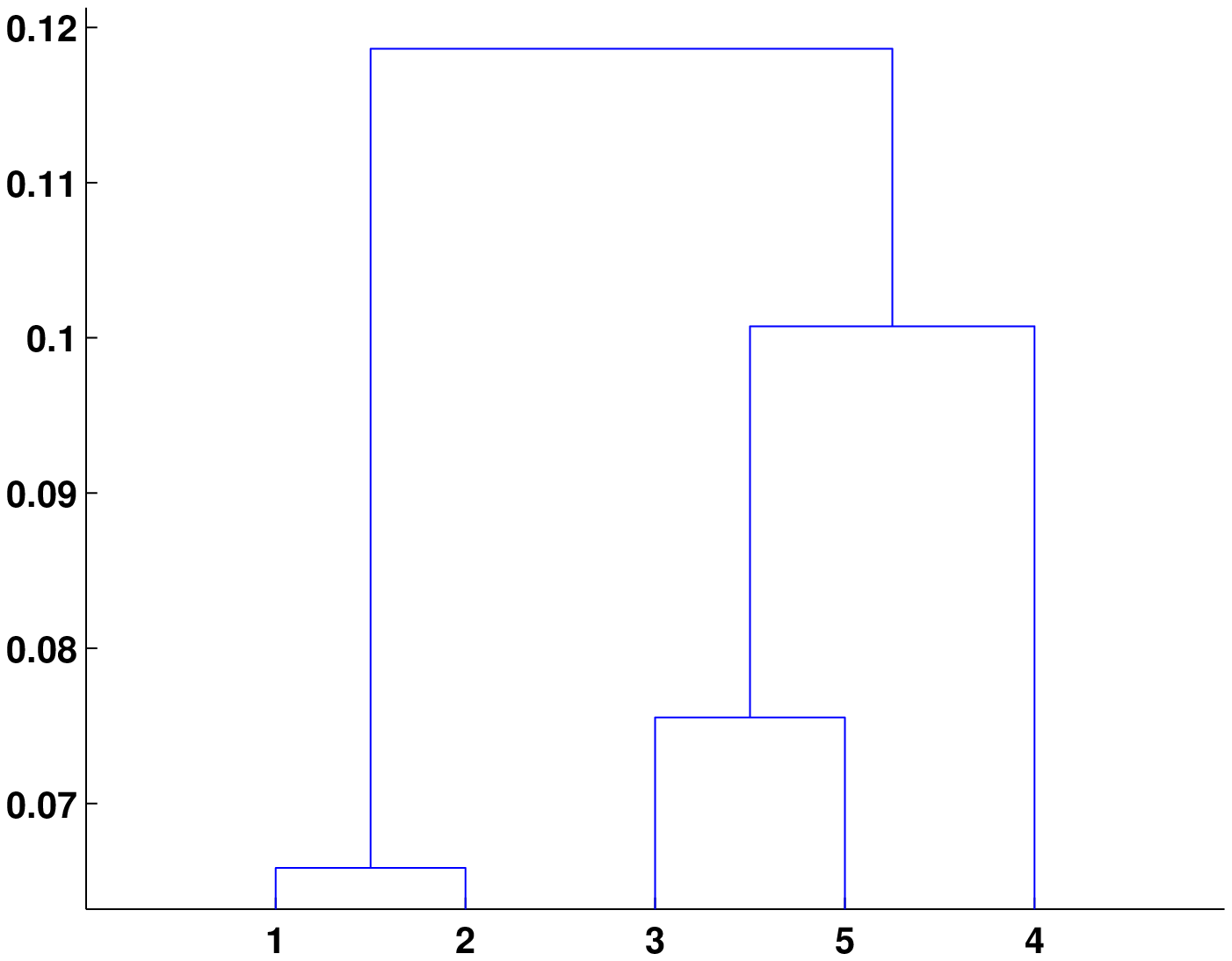}
       \label{fig:}
   }       
   \caption{\small{Hierarchical factor modeling results. (a) Factor loadings for E-coli data. (b) Inferred hierarchy for E-coli data. (c) Factor loadings for breast-cancer data. (d) Inferred hierarchy for breast-cancer data..}}
   \label{fig:hierfac}
\end{figure*}

\begin{figure}[t]
\includegraphics[width=1.9in]{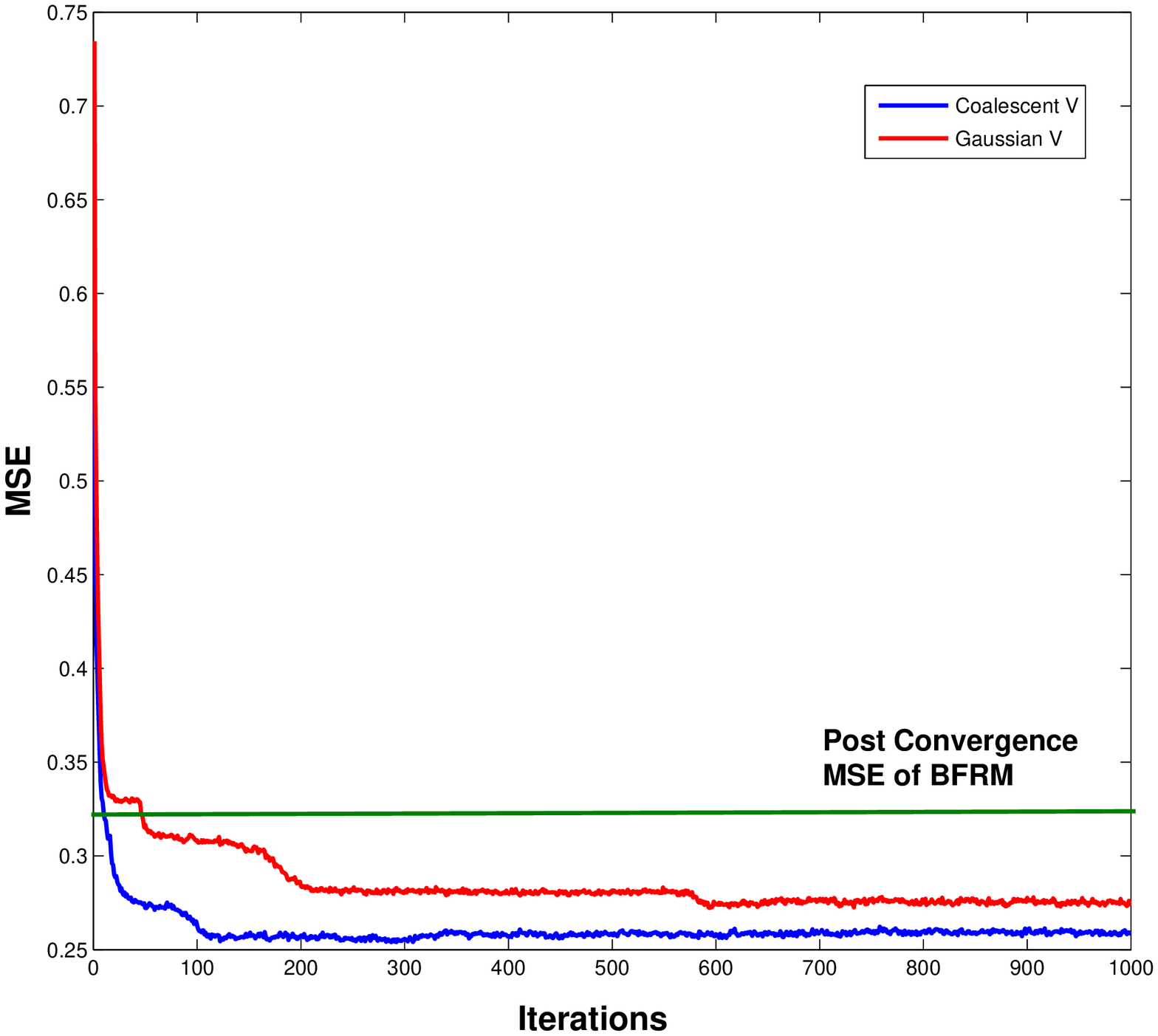}
\hspace{-1.8em}
\includegraphics[width=1.9in]{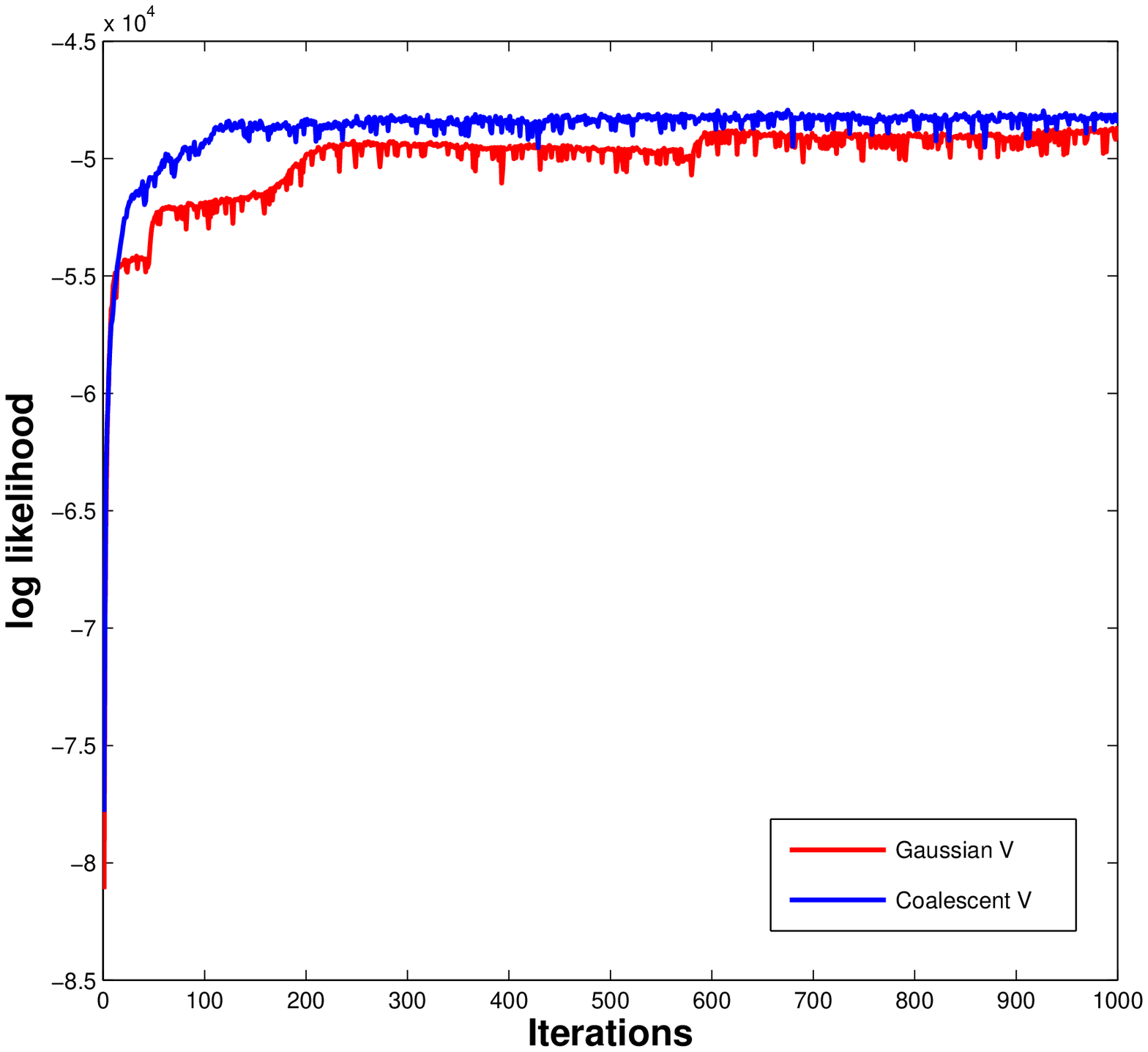}
\begin{minipage}[c]{0.8in}
\vspace{-2in}
\begin{small}
\begin{tabular}{@{}l@{~}|@{~}c@{~}|@{~}c@{}}
{\bf Model} & {\bf Binary}    & {\bf Real}\\
            & {\bf (\%error,std dev)} & {\bf (MSE)}\\
\hline
\textsf{LogReg}  & 17.5 (1.6) & - \\
\textsf{BFRM}  & 19.8 (1.4) & 0.48 \\
\textsf{Nor-V} & 15.8 (0.56) & 0.45 \\
\textsf{Coal-V} & 14.6 (0.48) & 0.43 \\
\textsf{PredModel} & 18.1 (2.1) & - \\
\end{tabular}
\end{small}
\end{minipage}
\vspace{-1em}
   \caption{\small{(a) MSE on the breast-cancer data for BFRM (horizontal line), our model with Gaussian (top red curved line) and Coalescent (bottom blue curved line) priors. This MSE is the reconstruction error for the data - different from the MSE for the held-out real valued responses (fig 7 c) (b) Log-likelihoods for our model with Gaussian (bottom red curved line) and Coalescent (top blue curved line) priors. (c) Factor regression results}}
\label{fig:coalres}
\end{figure}

Our results with hierarchical factor modeling are shown in figure \ref{fig:hierfac} for synthetic and real data. As shown, the model correctly infers the gene-factor associations, the number of factors, and the factor hierarchy. There are several ways to interpret the hierarchy. From the factor hierarchy for E-coli data (figure \ref{fig:hierfac}), we see that column-2 (corresponding to factor-2) of the \textbf{V} matrix is the most prominent one (it regulates the highest number of genes), and is closest to the tree-root, followed by column-2, which it looks most similar to. Columns corresponding to lesser prominent factors are located further down in the hierarchy (with appropriate relatedness). Figure \ref{fig:hierfac} (d) can be interpreted in a similar manner for breast-cancer data. The hierarchy can be used to find factors in order of their prominence. The higher we chop off the tree along the hierarchy, the more prominent the factors, we discover, are. For instance, if we are only interested in top 2 factors in E-coli data, we can chop off the tree above the sixth coalescent point. This is akin to the agglomerative clustering sense which is usually done \emph{post-hoc}. In contrast, our model discovers the factor hierarchies as part of the inference procedure itself. At the same time, there is no degradation of data reconstruction (in mean squared error sense) and the log-likelihood, when compared to the case with Gaussian prior on \textbf{V} (see figure \ref{fig:coalres} - they actually \emph{improve}). We also show in section \ref{sec:factreg} that hierarchical modeling results in better predictive performance for the factor regression task. Empirical evidences also suggest that the factor hierarchy leads to faster convergence since most of the unlikely configurations will never be visited as they are constrained by the hierarchy.

\begin{comment} 
\begin{figure*}[!htbp]
   \centering
   \subfigure[] {
       \includegraphics[width=2in]{mseplot.eps}
       \label{fig:}
   }
   \hspace{-2em}
   \subfigure[] {
       \includegraphics[width=2in]{loglikplot.eps}
       \label{fig:}
   }     
   \caption{\small{Gaussian vs Coalescent modeling of V matrix: (a) Mean squared errors on breast-cancer dataset (also shown is the post-convergence MSE of BFRM). (b) Log-likelihoods.}}
   \label{fig:comphier}
\end{figure*} 
\end{comment}

\vspace{-1em}
\subsection{Factor Regression}
\label{sec:factreg}
We report factor regression results for binary and real-valued responses and compare both variants of our model (Gaussian \textbf{V} and Coalescent \textbf{V}) against 3 different approaches: logistic regression, BFRM, and fitting a separate predictive model on the discovered factors (see figure \ref{fig:coalres} (c)). The breast-cancer dataset had two binary response variables (phenotypes) associated with each sample. For this binary prediction task, we split the data into training-set of 151 samples and test-set of 100 samples. This is essentially a transduction setting as described in section \ref{sec:regression} and shown in figure \ref{fig:lumped}. For real-valued prediction task, we treated a 30x20 block of the data matrix as our held-out data and predicted it based on the rest of the entries in the matrix. This method of evaluation is akin to the task of image reconstruction \cite{nlccapca}. The results are averaged over 20 random initializations and the low error variances suggest that our method is fairly robust w.r.t. initializations.

\section{Conclusions and Discussion}
\label{sec:concl}
\vspace{-1mm}
We have presented a fully nonparametric Bayesian approach to sparse
factor regression, modeling the gene-factor relationship using a
sparse variant of the IBP. 
%This leads to simple and efficient inference algorithms 
%based on sampling, obviating the need for model search. 
However, the true power of nonparametric
priors is evidenced by the ease of integration of task-specific models
into the framework.  Both gene selection and hierarchical factor
modeling are straightforward extensions in our model that do not
significantly complicate the inference procedure, but lead to improved
model performance \emph{and} more understandable outputs.  We applied
Kingman's coalescent as a hierarhical model on \textbf{V}, the matrix
modulating the expression levels of genes in factors.  An interesting
open question is whether the IBP can, itself, be modeled
hierarchically.

\vspace{-1.2em}
\small
\bibliographystyle{unsrt}
\bibliography{ihfrm}

\end{document}